\documentclass[10pt,twocolumn,letterpaper]{article}

\usepackage[pagenumbers]{cvpr} 

\usepackage[accsupp]{axessibility}
\definecolor{cvprblue}{rgb}{0.21,0.49,0.74}
\usepackage[pagebackref,breaklinks,colorlinks,allcolors=cvprblue]{hyperref}
\usepackage{amsmath}

\usepackage{amsfonts}
\usepackage{verbatim}
\usepackage{multirow}
\usepackage[table,xcdraw]{xcolor}
\usepackage{arydshln} 
\usepackage{pifont}
\usepackage{booktabs}
\newcommand{\cmark}{\ding{51}}%
\newcommand{\xmark}{\ding{55}}%

\usepackage{xcolor}

\def\paperID{37185} 
\def\confName{CVPR}
\def\confYear{2026}

\title{From Adaptation to Generalization: Adaptive Visual Prompting\\for Medical Image Segmentation}

\author{Evren Çetinkaya$^{1,}$\thanks{Equal Contribution} \quad Sangmin Lee$^{2,}$\footnotemark[\value{footnote}] \quad Jung Uk Kim$^3$ \quad Hong Joo Lee$^{4,}$\thanks{Corresponding Author:hongjoo.lee@seoultech.ac.kr} \quad Nassir Navab$^1$\\[3mm] 
{$^1$Technical University of Munich\quad$^2$Korea University}\quad$^3$Kyung Hee University\\[0.1mm]
{$^4$Seoul National University of Science and Technology}\\
{\tt\small \{evren.cetinkaya, nassir.navab\}@tum.de}\quad{\tt\small sangmin-lee@korea.ac.kr}\\ \quad{\tt\small ju.kim@khu.ac.kr}\quad{\tt\small hongjoo.lee@seoultech.ac.kr}
}

\begin{document}
\maketitle
\begin{abstract}
    Visual prompting has emerged as a powerful method for adapting pre-trained models to new domains without updating model parameters. However, existing prompting methods typically optimize a single prompt per domain and apply it uniformly to all inputs, limiting their ability to generalize under intra and inter-domain variability, which is especially critical in the medical field. To address this, we propose APEX, an \textbf{A}daptive \textbf{P}rompt \textbf{EX}traction framework that retrieves input-specific prompts from a learnable prompt memory. The memory stores diverse, domain-discriminative prompt representations and is queried via domain features extracted from the Fourier spectrum. To learn robust and discriminative domain features, we introduce a novel Low-Frequency Feature Contrastive (LFC) learning framework that clusters representations from the same domain while separating those from different domains. Extensive experiments on two medical segmentation tasks demonstrate that APEX significantly improves generalization across both seen and unseen domains. Furthermore, it complements any existing backbones and consistently enhances performance, confirming its effectiveness as a plug-and-play prompting solution in medical fields. The code is available at \url{https://github.com/cetinkayaevren/apex/}
\end{abstract}   

\section{Introduction}

Medical image segmentation is essential for computer-aided diagnosis \cite{seg_1,seg_2,lee2020structure,park2019endometrium}. Although Deep Neural Networks (DNNs) have significantly enhanced segmentation performance, their effectiveness often deteriorates when employed in clinical environments due to domain shift \cite{shift_1,shift_2}. Specifically, while models usually perform well on data from the trained domain (i.e., source domain), they often fail to generalize effectively to new domains (i.e., target domain). In the medical field, domain shifts frequently occur due to differences in imaging devices or image acquisition settings. To address this problem, numerous studies have explored domain adaptation methods, typically by fine-tuning models on target-domain data \cite{train_1,train_2,train_3}. However, these methods often suffer from catastrophic forgetting, where the model loses previously learned knowledge during adaptation and does not work properly on the source domain after adaptation \cite{cata_1,cata_2}. While it can be mitigated by retaining access to source domain data, this is often impractical in medical applications due to strict privacy regulations and data access constraints \cite{data_1,data_2}.

Recently, Visual Prompt (VP)-based domain adaptation methods have emerged as promising alternatives to alleviate these limitations \cite{vpt,vptta,vpad,fvp,a2xp,lee2026unsupervised}. Instead of modifying the pre-trained model parameters, VP methods add learnable parameters into the image space and optimize these parameters to reduce the loss for the target task. Since the original model parameters remain unchanged, VP methods inherently preserve source domain knowledge, substantially reducing catastrophic forgetting. Moreover, since VP optimization requires only target domain data, it alleviates privacy concerns associated with accessing source data. 

\begin{figure}[t]
	\centering
	    \includegraphics[width=0.99\linewidth]{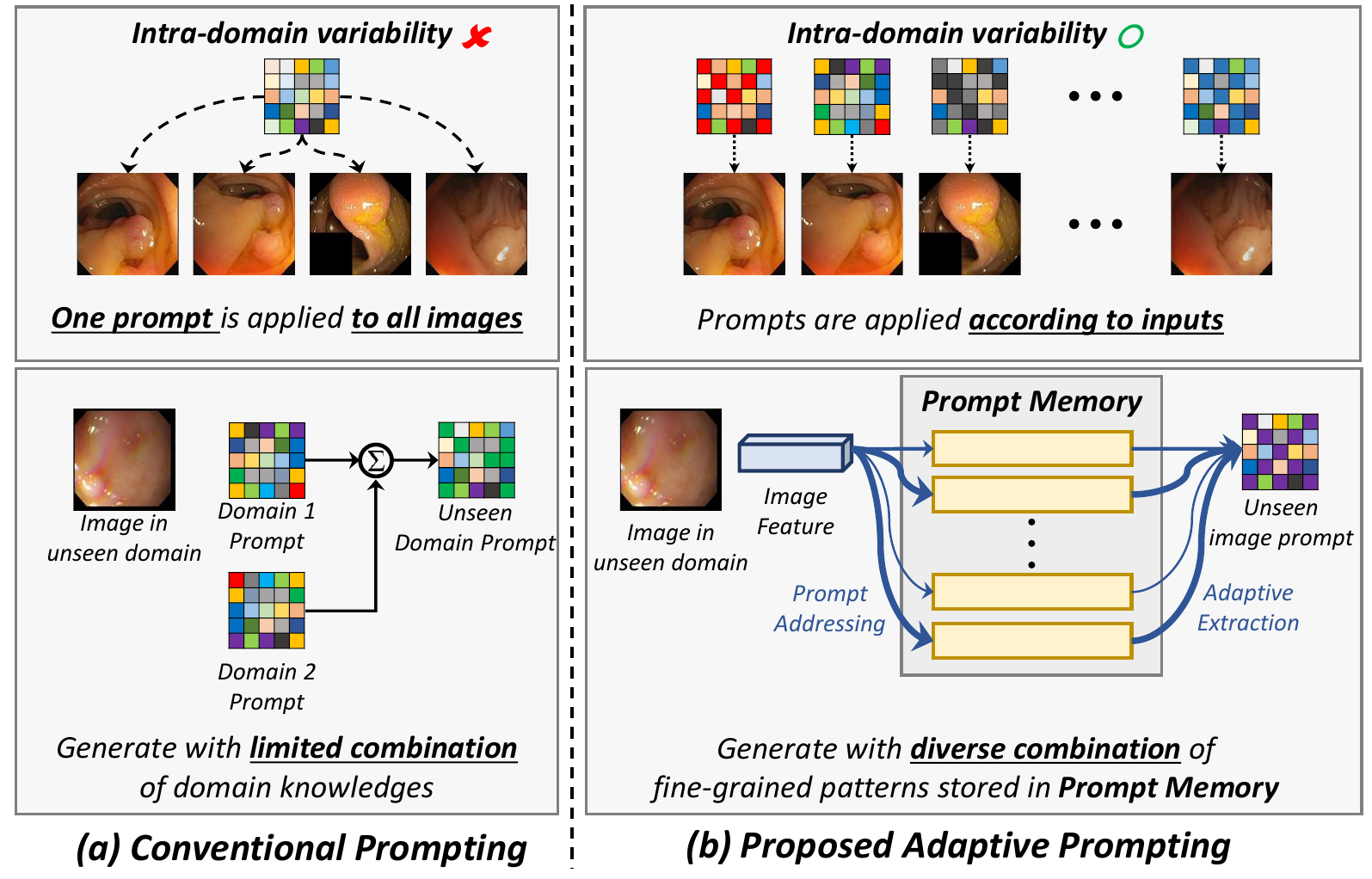}

	\caption{Graphical comparison with (a) conventional prompting approaches and (b) proposed adaptive prompting.} 
\label{fig:concept}
\vspace{-0.5cm}
\end{figure}

While these approaches have demonstrated promising domain adaptation capabilities, they typically optimize a single prompt for each target domain and apply the same prompt to all input images. This one-size-fits-all strategy fails to account for intra-domain variability, which refers to the heterogeneity among images within the same domain. For instance, medical images acquired from the same device can still differ significantly due to variations in acquisition settings or pathological features. Therefore, a single domain-level VP is too coarse to capture these fine-grained variations, leading to poor adaptation and clinical inaccuracy. Furthermore, these methods overlook inter-domain variability, which encompasses broader differences across domains, such as those arising from different imaging devices or clinical institutions. Relying on a fixed prompt optimized for a specific domain limits the ability of the pre-trained model to adapt to such diverse and complex input conditions. Moreover, existing approaches often rely on a limited set of coarse, domain-level prompts, which constrains their expressiveness and limits their ability to capture novel features not seen during training. Consequently, their generalization to unseen domains remains limited. This limitation is illustrated in Fig.~\ref{fig:concept} (a).

In this study, we propose a novel framework that addresses the limitations of existing VP methods in handling intra- and inter-domain variability. Unlike existing visual prompting methods that rely on a limited set of domain-level prompts, our approach adaptively generates input-specific prompts based on image features, enabling more flexible and fine-grained adaptation. As illustrated in Fig.~\ref{fig:concept} (a), conventional methods apply a single, static prompt identically to all images within a domain, ignoring the inherent intra-domain variability. In contrast, as shown in Fig.~\ref{fig:concept} (b), our framework adaptively applies a prompt conditioned on the individual input, which enables the model to capture intra-variability.

To this end, we introduce the \textbf{A}daptive \textbf{P}rompt \textbf{EX}traction (APEX) framework, which comprises a domain feature encoder, a prompt memory, and a prompt decoder. The domain feature encoder extracts discriminative features that can capture intra-domain variability in addition to inter-domain one. The prompt memory stores fine-grained prompt patterns corresponding to various images from target domains, allowing the framework to adaptively retrieve and combine the most relevant prompts based on the extracted domain features. This enables flexible combinations of existing knowledge at the feature level, significantly increasing the generalizability to diverse inputs, including unseen domains (See Fig. \ref{fig:concept}). 
Then, the resulting prompt feature is decoded into a final VP via the prompt decoder.

Furthermore, we propose a novel learning strategy to enhance domain discrimination in the domain feature encoder and promote diversity of prompt representations in the prompt memory. Specifically, we introduce Low-frequency Feature Contrastive (LFC) learning considering that low-frequency components in medical images vary significantly across domains~\cite{qiao2024medical,nam2024modality}. Through this learning, the domain feature encoder is optimized to encourage discriminating features from different domains, 
focusing specifically on the low-frequency spectrum. As a result, the domain feature encoder learns to capture inter-domain differences more effectively, enabling accurate retrieval of prompts from the memory bank across diverse domains.
The major contributions of our paper are as follows:
\begin{itemize}
    \item  We highlight key limitations of existing VP methods that cannot handle intra-domain variability and limited coarse combinations of domain-level prompts, which are crucial in the medical field.
    \item We propose APEX, an adaptive memory-based framework that flexibly retrieves input-specific prompts to handle intra-domain variability.
    \item We introduce a novel LFC learning that explicitly enhances inter-domain feature discrimination, promoting diversity in the prompt memory.
    \item By leveraging a prompt memory that stores diverse, domain-discriminative prompt representations, our framework significantly enhances generalization to not only seen domains but also unseen domains without further optimization or adaptation.
\end{itemize}

\begin{figure*}[t]
	\centering
	    \includegraphics[width=0.75\linewidth]{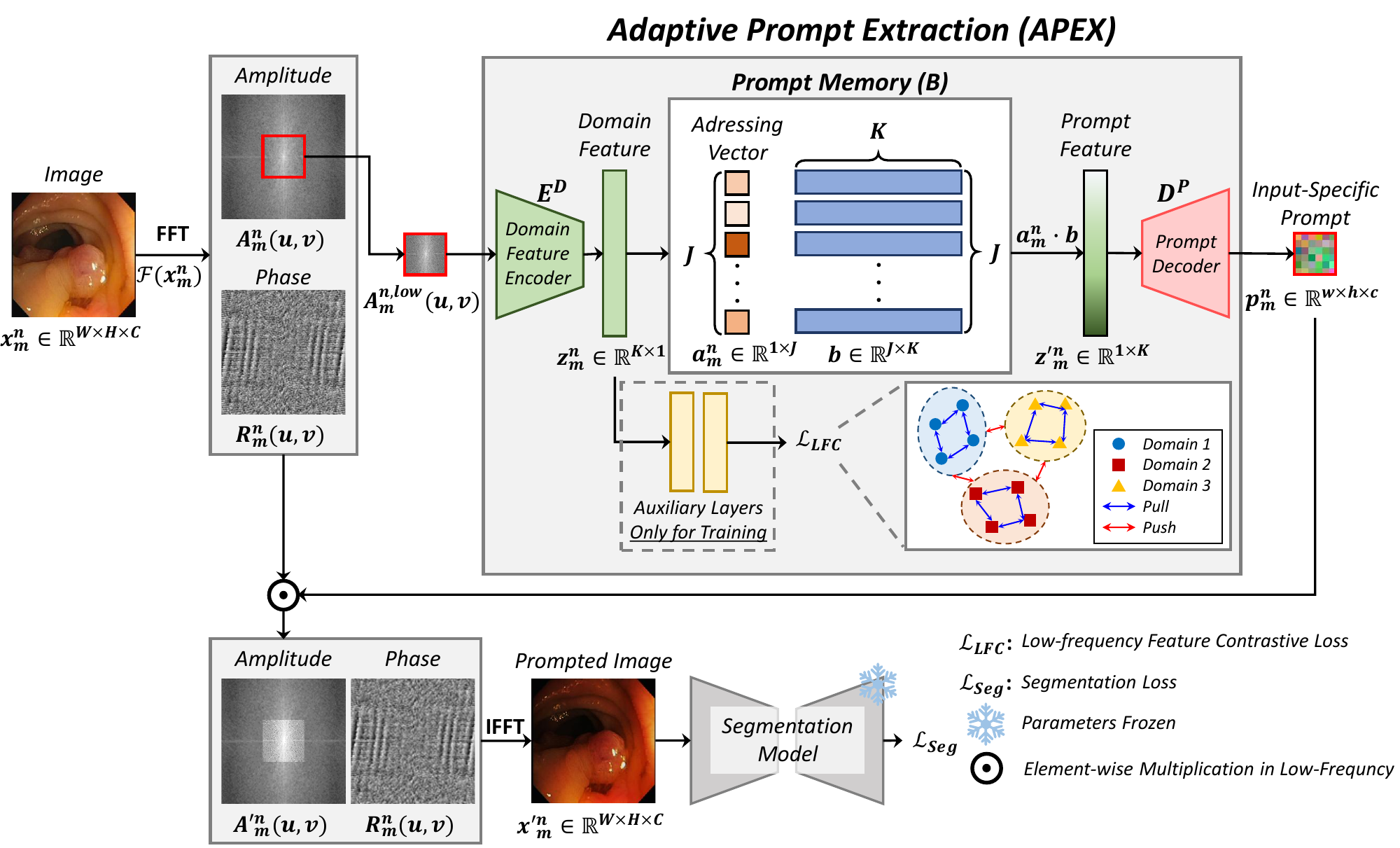}
	\caption{Overview of the proposed adaptive prompting framework. During optimization of APEX, the parameters of the segmentation model are frozen and only APEX is optimized.} 
\label{fig:method}
\vspace{-0.5cm}
\end{figure*}

\section{Related Work}
\subsection{Domain Adaptation in Medical Segmentation}
To address the domain shift problem in medical domain, a wide range of domain adaptation methods have been proposed \cite{hamed2019domain,conjeti2016supervised,yin2023class,huai2023context,yu2023source,tang2023source,goodman2023supervised,zhang2024mapseg,cui2024towards}. Traditional approaches typically involve fine-tuning the model parameters on target domain data, either through supervised \cite{hamed2019domain,conjeti2016supervised,goodman2023supervised} or unsupervised \cite{yu2023source,tang2023source,cai2025style} learning frameworks. Tang et al. \cite{tang2023source} exploited weakly augmented images to generate pseudo labels for a teacher network, while strong augmentations are applied to train the student network. Cai et al. \cite{cai2025style} presented Style Mixup Enhanced Disentanglement Learning (SMEDL), which injects “style‑mixed” feature representations to broaden style diversity and enable unsupervised domain adaptation. 

Although these methods have shown strong performance in domain adaptation, they are prone to catastrophic forgetting, where previously learned knowledge is forgotten during adaptation~\cite{cata_1,cata_2}. This can be mitigated by accessing the source domain data, but it is often infeasible in medical applications due to privacy and regulatory constraints~\cite{data_1}.

\subsection{Visual Prompting for Domain Adaptation}
Prompt learning was initially designed to provide additional text context for the input text to tune the Natural Language Processing (NLP) models or multi-modal models \cite{liu2023pre,Gunduboina_2025_CVPR}. After the success in NLP task, it is deployed in vision tasks. For the vision task, user-defined interactive VPs also provide additional context to the vision models as bounding box, points, etc \cite{kirillov2023segment,wu2024one,hu2024prompting}.

Different from user-defined prompts, learnable VPs can also be used for domain adaptation \cite{vpt,vpad,fvp,a2xp,vptta}. By optimizing learnable parameters added to input images, it projects input data to a new space and makes the input perform well on the pre-trained model. Since it does not change the parameters of the pre-trained model, it preserves source-domain knowledge and mitigates catastrophic forgetting, while avoiding direct access to source data. Thus, it can successfully adapt target domain data to pre-trained models.
Early work by Jia \textit{et al.} \cite{vpt} introduced VPs to adapt vision transformers. Meta-Padding \cite{vpad} extended this idea by inserting learnable padding tokens at the input and feature levels. FVP \cite{fvp} embedded prompts in the Fourier amplitude spectrum, and A2XP \cite{a2xp} unified domain generalization and adaptation in a single VP framework. VPTTA \cite{vptta} used the visual prompt for test-time adaptation.

Most of these methods optimize a single prompt per domain and apply it identically to every image, limiting their ability to handle intra-domain variability. Furthermore, although VPTTA adopted a memory structure, that memory is explicitly designed for the test-time adaptation setting, where the problem setting is different from ours. Our work tackles these limitations by introducing an adaptive prompt-extraction framework that retrieves input-specific prompts from memory. Through the input-specific prompt strategy, we could handle the intra-domain variability. Furthermore, by learning domain-discriminative features and storing diverse prompt representations in the prompt memory, we could capture the inter-domain variability. Therefore, we could make the prompt generalizable not only to seen domains (adapted domains) but also to unseen domains. 

\section{Proposed Method}
\subsection{Overview of Proposed Framework}
Fig. \ref{fig:method} shows an overview of the proposed prompting framework, which applies visual prompts adaptively based on input images. Given $N$ different domain datasets, each dataset $X^{n}$ consists of $M$ images, where $X^n=\{x^{n}_{1},x^{n}_{2},..., x^{n}_{M}\}$. Note that ${n}$ indicates a domain index while $m$ represents an image index within a domain. Each input image $x^{n}_{m}$ is transformed into the frequency domain using the Fast Fourier Transform (FFT), yielding its amplitude and phase components: ${A^{n}_{m}(u,v), R^{n}_{m}(u,v)} = \mathcal{F}(x^{n}_{m})$, where $\mathcal{F}(\cdot)$ denotes the FFT operation, $A^{n}_{m}(u,v)$ and $R^{n}_{m}(u,v)$ represent the amplitude and phase of the frequency component at location $(u,v)$, respectively. Then, we extract the low-frequency component $A^{n,low}_{m}(u,v)$ and feed it into the Adaptive Prompt EXtraction (APEX), which adaptively extracts a prompt. The extracted prompt is applied to the original amplitude, and the modified image is reconstructed via Inverse FFT (IFFT) before being fed into the segmentation model. 

\noindent\textbf{Prompting in Low-frequency Components:} In medical imaging, domain shifts often stem from differences in imaging devices or scanner settings, etc. These variations are closely associated with global appearance changes such as contrast, brightness, and shading, which are primarily encoded in the low-frequency components of the image spectrum. Prior studies have shown that such low-frequency variations contribute significantly to domain discrepancies~\cite{qiao2024medical,nam2024modality}. In contrast, high-frequency components and phase information capture fine anatomical details and structural layouts, which are essential for accurate segmentation. Altering them may distort spatial integrity and degrade performance. Therefore, we apply prompting in the low-frequency amplitude space to enable effective domain adaptation while preserving anatomical structure.

\subsection{Adaptive Prompt Extraction Module}
APEX module consists of three components: Domain Feature Encoder $E^{D}(\cdot)$, which extracts domain discriminative features from the low-frequency input; Prompt Memory ($B\in \mathbb{R}^{J\times K}$), which stores a set of $J$ learnable prompt vectors of dimension $K$; and Prompt Decoder $D^{P}(\cdot)$, which transforms the retrieved prompt features into a final prompt. 

The domain feature encoder extracts $K$-dimensional domain feature vector ($z^n_{m}\in\mathbb{R}^{K\times 1}$) from $E^{D}(A^{n,low}_{m}(u,v))$. Then, we retrieve the most suitable prompt feature from the memory $B$. $B$ consists of $J$ slots, each storing $K$-dimensional prompt vectors that capture diverse image properties, including both inter-domain and intra-domain differences. Also, we initialize the memory slots using orthogonal initialization to promote diversity and reduce redundancy among the slots. Then, to extract the final prompt feature $z'^n_{m}$, we compute the cosine similarity between $z^n_{m}$ and each stored prompt vector $b_j$, generating an addressing vector $a^n_{m}$. Each component in $a^n_{m}$ determines the contribution of each prompt vector for $z'^n_{m}$.
\begin{equation}
    a^n_{m,j}=\frac{z^n_m\cdot b_j}{\left\|z^n_m \right\|\left\|b_j \right\|}, \quad j\in\{1,2,...,J\}.
\end{equation}
With the calculated addressing vector, the final prompt feature $z'^n_m$ is extracted with a weighted sum over $B$. This can be written as follows:

\begin{equation}
    z'^n_m=\sum_{j=1}^{J}a^n_{m,j}\cdot b_j, \quad z'^n_m \in \mathbb{R}^{1 \times K}.
\end{equation}
The weights of $B$ are updated via backpropagation as \cite{lee2021video,lee2022weakly}. Since $z'^n_m$ is extracted by considering the contribution of each memory component, we can extract the most optimal prompt feature. Finally, $z'^n_m$ is decoded with $D^{P}(z'^n_m)$ and generates the prompt $p^n_m\in\mathbb{R}^{w\times h\times c}$. Then, $p^n_m$ is element-wise multiplied with $A^{n,low}_m$ to apply the prompt, and IFFT is applied to get the prompted image $x'^n_m$.

\begin{figure}[t]
	\centering
	    \includegraphics[width=0.9\linewidth]{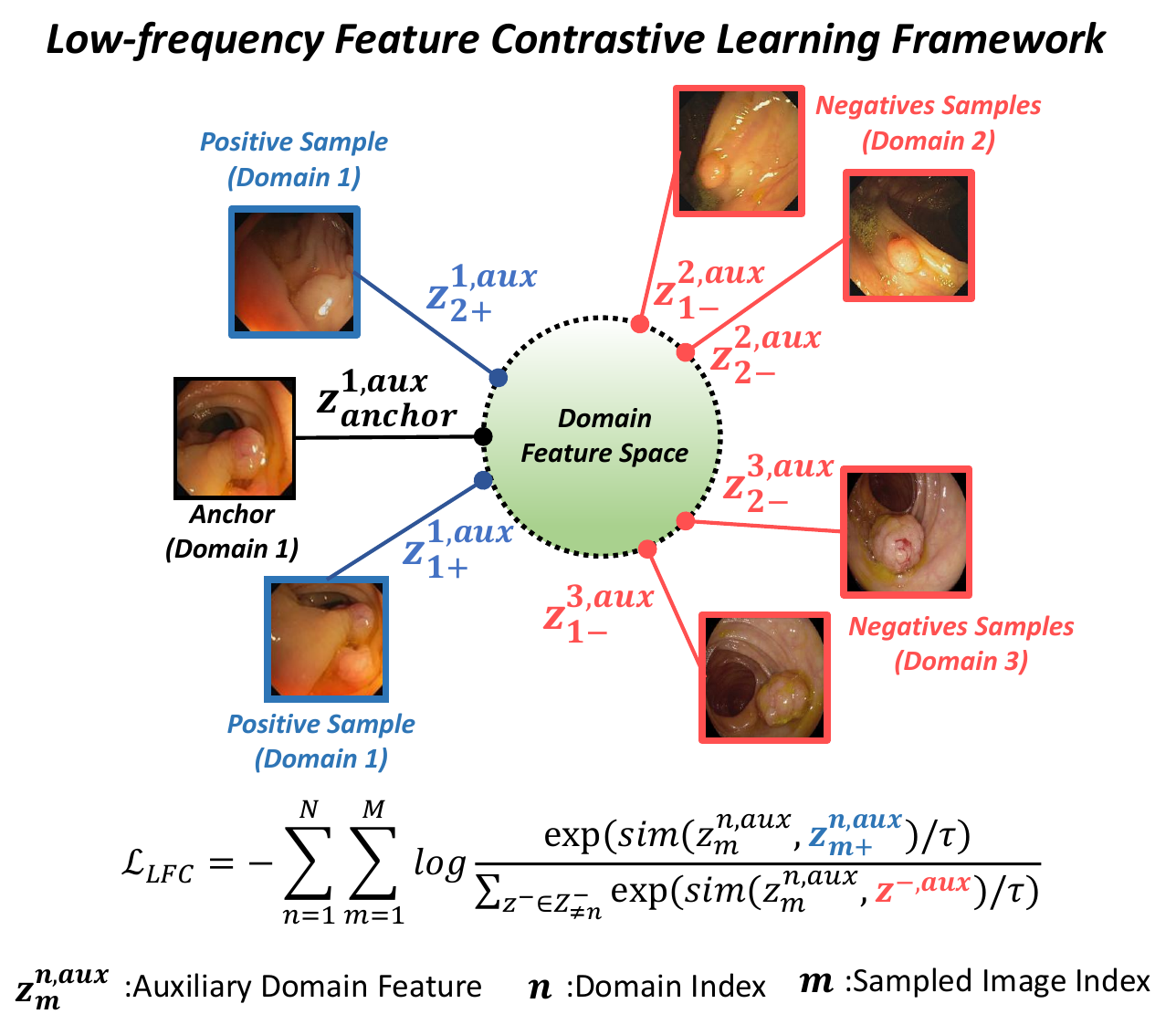}
        \vspace{-0.3cm}
	\caption{Visual descriptions about the LFC learning framework.} 
\label{fig:lfc}
\vspace{-0.3cm}
\end{figure}

\subsection{Low-frequency Feature Contrastive Learning}
To train APEX effectively, we employ two loss functions: a segmentation loss $\mathcal{L}_{\text{Seg}}$ and a proposed Low-frequency Feature Contrastive loss $\mathcal{L}_{\text{LFC}}$. The segmentation loss is a conventional loss with Dice and Cross-Entropy losses. To further guide the domain encoder in learning both inter-domain and intra-domain discriminative features, we introduce $\mathcal{L}_{\text{LFC}}$, which operates on low-frequency features known to carry domain-specific information~\cite{qiao2024medical,nam2024modality}.

Motivated by this, $\mathcal{L}_{\text{LFC}}$ enforces that low-frequency features from the same domain are pulled closer in the feature space, while low-frequency features from different domains are pushed apart. This encourages domain-aware clustering while preserving intra-domain discrimination. To apply this loss, we use an auxiliary projection head (two MLP layers) on top of the domain feature $z^n_m$ to avoid feature collapse during optimizing $E^D$, producing an auxiliary embedding $z^{n,\text{aux}}_m$. This projection is used only for training and discarded at inference time. It can be formulated as:
\begin{equation}
    \mathcal{L}_{\text{LFC}} = -\sum_{n=1}^{N} \sum_{m=1}^{M} \log \frac{\exp\left( \text{sim}(z^{n,\text{aux}}_m, z^{n,\text{aux}}_{m^+}) / \tau \right)}
    {\sum\limits_{z^- \in Z^{-}_{\neq n}} \exp\left( \text{sim}(z^{n,\text{aux}}_m, z^{-,\text{aux}}) / \tau \right)},
\end{equation}
where $z^{n,\text{aux}}_{m^+}$ is a positive pair sampled from the same domain $n$ (with $m \neq m^+$), $Z^{-}_{\neq n}$ denotes negative samples from all other domains, $\text{sim}(a, b)$ is cosine similarity, and $\tau$ is a temperature scaling parameter that effectively scaling the similarity scores to balance the emphasis between positive and negative pairs during training. By clustering features from the same domain, the proposed LFC learning framework enables the $E^D$ to capture both the shared characteristics of the domain and the variations within it. Simultaneously, by separating features from different domains, $E^D$ learns domain-discriminative features while also capturing inter-domain variability. We graphically describe the LFC learning framework in Fig.~\ref{fig:lfc}

\begin{table*}[!t]
\centering
\renewcommand{\arraystretch}{1.1}
\resizebox{0.77\textwidth}{!}{%
\begin{tabular}{clcccccccccc}
\specialrule{.1em}{.0em}{0.em}
                                     & \multicolumn{1}{c}{}                                  & \multicolumn{4}{c}{\textbf{Seen}}                                                                                                                                 & \multicolumn{4}{c}{\textbf{Unseen}}                                                                                                                               & \multicolumn{2}{c}{}                                                            \\ \cline{3-10}
                                     & \multicolumn{1}{c}{}                                  & \multicolumn{2}{c}{\textbf{Domain A}}                                           & \multicolumn{2}{c}{\textbf{Domain B}}                                           & \multicolumn{2}{c}{\textbf{Domain C}}                                           & \multicolumn{2}{c}{\textbf{Domain D}}                                           & \multicolumn{2}{c}{\multirow{-2}{*}{\textbf{Avg}}}                              \\ \cline{3-12} 
\multirow{-3}{*}{\textbf{Backbone}}  & \multicolumn{1}{c}{\multirow{-3}{*}{\textbf{Method}}} & \textbf{DICE}                          & \textbf{IoU}                           & \textbf{DICE}                          & \textbf{IoU}                           & \textbf{DICE}                          & \textbf{IoU}                           & \textbf{DICE}                          & \textbf{IoU}                           & \textbf{DICE}                          & \textbf{IoU}                           \\ \hline
                                     & Source Only                                                  & 76.67                                  & 69.25                                  & 86.20                                  & 79.04                                  & 51.64                                  & 45.71                                  & 57.52                                  & 52.01                                  & 65.34                                  & 61.36                                  \\ \cdashline{2-12}
                                     & VPT (ECCV'22)                                         & 76.82                                  & 70.07                         & 86.12                                  & 78.93                                  & 51.05                                  & 45.71                                  & 57.55                                  & 51.96                                  & 67.88                                  & 61.67                                  \\
                                     & FVP (TMI'23)                                          & 75.06                                  & 69.81                                  & 86.95                                  & 79.32                                  & 51.41                                  & 42.38                                  & 59.11                                  & 52.78                                  & 67.63                                  & 60.82                                  \\
                                     & A2XP (CVPR'24)                                        & 76.79                                  & 70.01                                  & 86.19                                  & 78.98                                  & 51.02                                  & 45.66                                  & 57.53                                  & 51.98                                  & 67.88                                  & 61.66                                  \\
                                     & VPAD (TPAMI'25)                                       & 77.46                                  & 70.91                                  & 87.43                                  & 79.31                                  & 52.44                                  & 46.75                                  & 58.10                                  & 52.81                                  & 67.45                                  & 61.15                                    \\
\multirow{-6}{*}{\textbf{PraNet}}    & \cellcolor[HTML]{EFEFEF}\textbf{APEX(Ours)}           & \cellcolor[HTML]{EFEFEF}\textbf{79.37} & \cellcolor[HTML]{EFEFEF}\textbf{72.20}          & \cellcolor[HTML]{EFEFEF}\textbf{88.13} & \cellcolor[HTML]{EFEFEF}\textbf{81.06} & \cellcolor[HTML]{EFEFEF}\textbf{54.43} & \cellcolor[HTML]{EFEFEF}\textbf{48.93} & \cellcolor[HTML]{EFEFEF}\textbf{60.85} & \cellcolor[HTML]{EFEFEF}\textbf{54.91} & \cellcolor[HTML]{EFEFEF}\textbf{70.95} & \cellcolor[HTML]{EFEFEF}\textbf{64.52} \\ \hline
                                     & Source Only                                                  & 63.53                                  & 53.65                                  & 76.10                                  & 66.37                                  & 44.79                                  & 35.07                                  & 52.79                                  & 44.04                                  & 60.38                                  & 51.16                                  \\ \cdashline{2-12}
                                     & VPT (ECCV'22)                                         & 63.22                                  & 53.33                                  & 75.77                                  & 66.08                                  & 44.48                                  & 34.79                                  & 52.60                                  & 43.86                                  & 60.12                                  & 50.87                                  \\
                                     & FVP (TMI'23)                                          & 64.85                                  & 54.38                                  & 76.82                                  & 66.87                                  & 48.04                                  & 37.89                                  & 53.09                                  & 43.32                                  & 60.20                                  & 50.29                                  \\
                                     & A2XP (CVPR'24)                                        & 63.53                                  & 53.65                                  & 76.10                                  & 66.38                                  & 44.80                                  & 35.07                                  & 52.80                                  & 44.04                                  & 60.38                                  & 51.18                                  \\
                                     & VPAD (TPAMI'25)                                        & 64.51                                  & 54.40                                  & 76.91                                  & 66.93                                  & 48.56                                  & 37.99                                  & 53.13                                  & 43.40                                  & 60.14                                  & 52.19                                    \\
\multirow{-6}{*}{\textbf{SwinUNet}}  & \cellcolor[HTML]{EFEFEF}APEX(Ours)                    & \cellcolor[HTML]{EFEFEF}\textbf{66.68} & \cellcolor[HTML]{EFEFEF}\textbf{56.43} & \cellcolor[HTML]{EFEFEF}\textbf{78.25} & \cellcolor[HTML]{EFEFEF}\textbf{68.38} & \cellcolor[HTML]{EFEFEF}\textbf{49.62} & \cellcolor[HTML]{EFEFEF}\textbf{39.97} & \cellcolor[HTML]{EFEFEF}\textbf{55.21} & \cellcolor[HTML]{EFEFEF}\textbf{45.76} & \cellcolor[HTML]{EFEFEF}\textbf{62.99} & \cellcolor[HTML]{EFEFEF}\textbf{53.28} \\ \hline
& Source Only                                                  & 73.02                                  & 64.54                                  & 75.94                                  & 67.67                                  & 59.30                                  & 53.03                                  & 66.31                                  & 58.69                                  & 69.63                                  & 61.17                                  \\ \cdashline{2-12}
                                     & VPT (ECCV'22)                                         & 73.11                                  & 64.66                                  & 75.83                                  & 67.58                                  & 59.54                                  & 53.37                                  & 66.18                                  & 58.58                                  & 69.57                                  & 61.73                                  \\
                                     & FVP (TMI'23)                                          & 71.64                                  & 62.48                                  & 79.10                                  & 70.90                                  & 64.80                                  & 56.57                                  & 66.56                                  & 58.51                                  & 70.67                                  & 62.37                                  \\
                                     & A2XP (CVPR'24)                                        & 73.03                                  & 64.54                                  & 75.94                                  & 67.67                                  & 59.32                                  & 53.04                                  & 66.31                                  & 58.69                                  & 69.63                                  & 61.77                                  \\
                                     & VPAD (TPAMI'25)                                       & 73.18                                  & 64.55                                  & 76.81                                  & 68.91                                  & 60.63                                  & 54.88                                  & 66.11                                  & 58.50                                  & 69.61                                  & 61.74                                    \\
\multirow{-6}{*}{\textbf{TransUNet}} & \cellcolor[HTML]{EFEFEF}APEX(Ours)                    & \cellcolor[HTML]{EFEFEF}\textbf{73.71} & \cellcolor[HTML]{EFEFEF}\textbf{64.87} & \cellcolor[HTML]{EFEFEF}\textbf{79.75} & \cellcolor[HTML]{EFEFEF}\textbf{71.86} & \cellcolor[HTML]{EFEFEF}\textbf{68.56} & \cellcolor[HTML]{EFEFEF}\textbf{60.20} & \cellcolor[HTML]{EFEFEF}\textbf{67.11} & \cellcolor[HTML]{EFEFEF}\textbf{59.96} & \cellcolor[HTML]{EFEFEF}\textbf{71.77} & \cellcolor[HTML]{EFEFEF}\textbf{63.49} \\ \specialrule{.1em}{.0em}{0.em}
\end{tabular}}
\caption{DICE (\%) and IoU (\%) comparison with recent VP methods across four backbone models on both seen and unseen domains for the polyp segmentation task. Domain A: CVC-ClinicDB, Domain B: Kvasir-Seg, Domain C: ETIS-LaribPolypDB, and Domain D: CVC-ColonDB. Source Only indicates the model before adaptation; even seen domains are treated as unseen in this case.}
\vspace{-0.3cm}
\label{tab:polyp}
\end{table*}

\subsection{Optimizing APEX}

APEX is optimized using two objectives: the segmentation loss \( \mathcal{L}_{\text{Seg}} \), and the Low-frequency Feature Contrastive loss \( \mathcal{L}_{\text{LFC}} \). The overall training objective is $
\mathcal{L}_{\text{total}} = \mathcal{L}_{\text{Seg}} + \mathcal{L}_{\text{LFC}}.
$
During training, \( \mathcal{L}_{\text{Seg}} \) is backpropagated through the decoder, prompt memory, and domain encoder. The prompt memory \( B \in \mathbb{R}^{J \times K} \) is updated via gradient descent. The gradient of \( \mathcal{L}_{\text{Seg}} \) with respect to \( B \) is formulated as:
\begin{equation}
\frac{\partial \mathcal{L}_{\mathrm{Seg}}}{\partial B} = \sum_{n,m}  {a^n_{m}} \ (\frac{\partial \mathcal{L}_{\mathrm{Seg}}}{\partial z'^n_m})^\top,
\end{equation}
where \({a^n_{m}}\) is the addressing vector. \( a^n_{m,j} \) is the attention weight assigned to the \( j \)-th memory slot for the \( m \)-th image in the \( n \)-th domain. Then, each memory slot \( b_j \in B \) is updated by:
\begin{equation}
b_j \gets b_j - \eta \cdot \frac{\partial \mathcal{L}_{\mathrm{Seg}}}{\partial b_j},
\end{equation}
where \( \eta \) is the learning rate and gradients are propagated only through the attention-weighted contributions. Then, $E^D$ is jointly optimized using \( \mathcal{L}_{\text{Seg}} \) and \( \mathcal{L}_{\text{LFC}} \).

\begin{table*}[!t]
\centering
\renewcommand{\arraystretch}{1.1}
\resizebox{0.77\textwidth}{!}{%
\begin{tabular}{clcccccccccc}
\specialrule{.1em}{.0em}{0.em}
                                     & \multicolumn{1}{c}{}                                  & \multicolumn{4}{c}{\textbf{Seen}}                                                                                                                                 & \multicolumn{4}{c}{\textbf{Unseen}}                                                                                                                               & \multicolumn{2}{c}{}                                                            \\ \cline{3-10}
                                     & \multicolumn{1}{c}{}                                  & \multicolumn{2}{c}{\textbf{Domain A}}                                           & \multicolumn{2}{c}{\textbf{Domain B}}                                           & \multicolumn{2}{c}{\textbf{Domain C}}                                           & \multicolumn{2}{c}{\textbf{Domain D}}                                           & \multicolumn{2}{c}{\multirow{-2}{*}{\textbf{Avg}}}                              \\ \cline{3-12} 
\multirow{-3}{*}{\textbf{Backbone}}  & \multicolumn{1}{c}{\multirow{-3}{*}{\textbf{Method}}} & \textbf{DICE}                          & \textbf{IoU}                           & \textbf{DICE}                          & \textbf{IoU}                           & \textbf{DICE}                          & \textbf{IoU}                           & \textbf{DICE}                          & \textbf{IoU}                           & \textbf{DICE}                          & \textbf{IoU}                           \\ \hline
                                     & Source Only                                                  & 84.73                                  & 76.16                                  & 81.65                                  & 72.33                                  & 67.58                                  & 58.68                                  & 37.74                                  & 34.39                                  & 73.71                                  & 65.51                                  \\ \cdashline{2-12}
                                     & VPT (ECCV'22)                                         & 85.40                                  & 77.36                                  & 82.57                                  & 73.39                                  & 68.38                                  & 59.68                                  & 38.73                                  & 35.66                                  & 74.17                                  & 66.23                                  \\
                                     & FVP (TMI'23)                                          & 87.19                                  & 79.07                                  & 84.00                                  & 75.13                                  & 73.46                                  & 64.03                                  & 39.32                                  & 35.48                                  & 76.42                                  & 68.40                                  \\
                                     & A2XP (CVPR'24)                                        & 85.43                                  & 76.41                                  & 82.87                                  & 73.51                                  & 68.44                                  & 59.46                                  & 38.53                                  & 35.87                                  & 74.15                                  & 66.21                                  \\
                                     & VPAD (TPAMI'25)                                       & 86.14                                  & 78.48                                  & 83.51                                  & 74.44                                  & 69.51                                  & 60.13                                  & 42.00                                  & 37.15                                  & 75.38                                  & 67.94                                  \\
\multirow{-6}{*}{\textbf{UNet}}      & \cellcolor[HTML]{EFEFEF}\textbf{APEX(Ours)}           & \cellcolor[HTML]{EFEFEF}\textbf{89.57} & \cellcolor[HTML]{EFEFEF}\textbf{82.42} & \cellcolor[HTML]{EFEFEF}\textbf{86.54} & \cellcolor[HTML]{EFEFEF}\textbf{78.28} & \cellcolor[HTML]{EFEFEF}\textbf{84.84} & \cellcolor[HTML]{EFEFEF}\textbf{76.67} & \cellcolor[HTML]{EFEFEF}\textbf{79.18} & \cellcolor[HTML]{EFEFEF}\textbf{69.47} & \cellcolor[HTML]{EFEFEF}\textbf{85.81} & \cellcolor[HTML]{EFEFEF}\textbf{77.56} \\ \hline
                                     & Source Only                                                  & 79.08                                  & 69.48                                  & 72.98                                  & 62.34                                  & 78.24                                  & 68.63                                  & 63.92                                  & 54.77                                  & 73.53                                  & 63.46                                  \\ \cdashline{2-12}
                                     & VPT (ECCV'22)                                         & 80.27                                  & 70.21                                  & 74.43                                  & 63.53                                  & 79.26                                  & 69.34                                  & 64.75                                  & 54.25                                  & 73.42                                  & 64.75                                  \\
                                     & FVP (TMI'23)                                          & 81.81                                  & 72.64                                  & 77.12                                  & 66.56                                  & 78.16                                  & 68.31                                  & 62.31                                  & 53.56                                  & 75.94                                  & 66.04                                  \\
                                     & A2XP (CVPR'24)                                        & 80.77                                  & 70.27                                  & 74.58                                  & 63.71                                  & 79.04                                  & 69.87                                  & 64.44                                  & 55.13                                  & 74.33                                  & 64.74                                  \\
                                     & VPAD (TPAMI'25)                                       & 81.31                                  & 71.33                                  & 76.41                                  & 65.10                                  & 79.10                                  & 60.23                                  & 64.55                                  & 54.79                                  & 74.99                                  & 65.84                                  \\
\multirow{-6}{*}{\textbf{ResUNet}}   & \cellcolor[HTML]{EFEFEF}\textbf{APEX(Ours)}           & \cellcolor[HTML]{EFEFEF}\textbf{88.59} & \cellcolor[HTML]{EFEFEF}\textbf{81.05} & \cellcolor[HTML]{EFEFEF}\textbf{80.95} & \cellcolor[HTML]{EFEFEF}\textbf{71.12} & \cellcolor[HTML]{EFEFEF}\textbf{79.35} & \cellcolor[HTML]{EFEFEF}\textbf{70.32} & \cellcolor[HTML]{EFEFEF}\textbf{75.17} & \cellcolor[HTML]{EFEFEF}\textbf{65.21} & \cellcolor[HTML]{EFEFEF}\textbf{81.42} & \cellcolor[HTML]{EFEFEF}\textbf{72.14} \\ \hline
                                     & Source Only                                                  & 85.31                                  & 76.23                                  & 86.85                                  & 78.67                                  & 87.34                                  & 79.58                                  & 79.54                                  & 69.52                                  & 84.76                                  & 76.00                                  \\ \cdashline{2-12}
                                     & VPT (ECCV'22)                                         & 86.36                                  & 77.62                                  & 87.22                                  & 79.09                                  & 87.26                                  & 79.37                                  & 79.97                                  & 69.97                                  & 85.20                                  & 76.51                                  \\
                                     & FVP (TMI'23)                                          & 85.99                                  & 77.12                                  & 86.08                                  & 77.53                                  & 86.01                                  & 77.75                                  & 79.11                                  & 69.12                                  & 84.30                                  & 75.38                                  \\
                                     & A2XP (CVPR'24)                                        & 86.03                                  & 77.20                                  & 87.21                                  & 79.22                                  & 87.24                                  & 79.34                                  & 79.72                                  & 69.70                                  & 85.05                                  & 76.37                                  \\
                                     & VPAD (TPAMI'25)                                       & 85.61                                  & 77.18                                  & 86.56                                  & 77.50                                  & 86.99                                  & 77.39                                  & 79.51                                  & 69.53                                  & 85.39                                  & 76.84                                  \\
\multirow{-6}{*}{\textbf{SwinUNet}}  & \cellcolor[HTML]{EFEFEF}\textbf{APEX(Ours)}           & \cellcolor[HTML]{EFEFEF}\textbf{89.09} & \cellcolor[HTML]{EFEFEF}\textbf{81.73} & \cellcolor[HTML]{EFEFEF}\textbf{87.26} & \cellcolor[HTML]{EFEFEF}\textbf{79.38} & \cellcolor[HTML]{EFEFEF}\textbf{88.30} & \cellcolor[HTML]{EFEFEF}\textbf{80.49} & \cellcolor[HTML]{EFEFEF}\textbf{80.67} & \cellcolor[HTML]{EFEFEF}\textbf{70.35} & \cellcolor[HTML]{EFEFEF}\textbf{86.58} & \cellcolor[HTML]{EFEFEF}\textbf{78.24} \\ \hline
& Source Only                                                  & 85.89                                  & 77.45                                  & 76.91                                  & 67.07                                  & 90.81                                  & 84.40                                  & 83.58                                  & 74.48                                  & 84.30                                  & 75.85                                  \\ \cdashline{2-12}
                                     & VPT (ECCV'22)                                         & 88.46                                  & 80.85                                  & 85.34                                  & 76.99                                  & 90.46                                  & 83.88                                  & 82.71                                  & 73.66                                  & 85.99                                  & 78.10                                  \\
                                     & FVP (TMI'23)                                          & 88.09                                  & 80.39                                  & 82.10                                  & 73.03                                  & 90.76                                  & 83.30                                  & 83.65                                  & 74.56                                  & 86.15                                  & 78.07                                  \\
                                     & A2XP (CVPR'24)                                        & 87.73                                  & 79.87                                  & 80.84                                  & 71.55                                  & 90.62                                  & 83.12                                  & 83.78                                  & 74.19                                  & 85.74                                  & 77.56                                  \\
                                     & VPAD (TPAMI'25)                                       & 88.00                                  & 80.47                                  & 81.98                                  & 71.74                                  & 90.11                                  & 83.89                                  & 83.19                                  & 74.08                                  & 85.18                                  & 77.95                                  \\
\multirow{-6}{*}{\textbf{TransUNet}} & \cellcolor[HTML]{EFEFEF}\textbf{APEX(Ours)}           & \cellcolor[HTML]{EFEFEF}\textbf{89.82} & \cellcolor[HTML]{EFEFEF}\textbf{82.75} & \cellcolor[HTML]{EFEFEF}\textbf{85.70} & \cellcolor[HTML]{EFEFEF}\textbf{77.43} & \cellcolor[HTML]{EFEFEF}\textbf{91.13} & \cellcolor[HTML]{EFEFEF}\textbf{85.05} & \cellcolor[HTML]{EFEFEF}\textbf{84.69} & \cellcolor[HTML]{EFEFEF}\textbf{75.63} & \cellcolor[HTML]{EFEFEF}\textbf{87.61} & \cellcolor[HTML]{EFEFEF}\textbf{79.58} \\ \specialrule{.1em}{.0em}{0.em}
\end{tabular}}
\caption{DICE (\%) and IoU (\%) comparison with recent VP methods across four backbone models on both seen and unseen domains for the OC/OD segmentation task. Domain A: REFUGE\_Test, Domain B: REFUGE\_Val, Domain C: Drishti-GS, and Domain D: RIM-ONE-r3. Source Only indicates the model before adaptation; even seen domains are treated as unseen in this case.}
\label{tab:optic}

\end{table*}

\section{Experiments}
\subsection{Experiments Settings}
\noindent\textbf{Dataset:}
We evaluate our method on two medical segmentation tasks: Polyp segmentation and Optic Cup/Disk (OC/OD) segmentation. For each task, we use five publicly available datasets, divided into training source models, prompt optimization, and evaluation sets across unseen domains.
For the polyp segmentation task, the segmentation model is trained on the BKAI \cite{bkai} dataset. Prompts are optimized using the training set of CVC-ClinicDB and Kvasir-Seg dataset, and evaluated with the test set of CVC-ClinicDB (Domain A) \cite{clinic} and Kvasir-Seg (Domain B) \cite{kvasir} for seen domains. To evaluate the generalizability to unseen domains, the test set of ETIS-LaribPolypDB (Domain C) \cite{etis} and CVC-ColonDB (Domain D) \cite{colon} were employed.

For OC/OD segmentation, we use the ORIGA dataset as the source domain to train the segmentation model. The prompts are optimized using the training set of the REFUGE dataset and evaluated on REFUGE\_Test (Domain A) and REFUGE\_Val (Domain B) \cite{orlando2020refuge} as seen domains. To evaluate generalization capability on unseen domains, we used test sets of the Drishti-GS (Domain C) \cite{sivaswamy2014drishti} and RIM-ONE-r3 (Domain D) \cite{fumero2011rim} datasets. Also, we used a domain augmentation method \cite{qiao2024medical} to generate synthetic domains. Details are presented in the supplementary material.

\noindent\textbf{Segmentation Backbone:} We evaluate our method using five segmentation backbones: UNet~\cite{ronneberger2015u}, ResUNet~\cite{zhang2018road}, SwinUNet~\cite{swinunet}, and TransUNet~\cite{chen2024transunet}, which are widely used in medical image segmentation, along with SOTA models for polyp segmentation known for strong generalization performance: PraNet~\cite{fan2020pranet}. The domain encoder and prompt decoder are implemented with four-layer multilayer perceptrons (MLPs). The number of memory slots in the prompt memory is fixed at $J=150$

\subsection{Quantitative Results}

\noindent\textbf{Polyp Segmentation:} To evaluate the effectiveness of the proposed method on the polyp segmentation task, we trained the APEX module using the training sets of CVC-ClinicDB and Kvasir-Seg. We then evaluated it on the test sets of CVC-ClinicDB (Domain A) and Kvasir-Seg (Domain B) as seen domains, and ETIS-LaribPolypDB (Domain C) and CVC-ColonDB (Domain D) as unseen domains. Table~\ref{tab:polyp} shows the Dice Similarity Coefficient (DICE) and Intersection over Union (IoU) scores for various SOTA VP-based adaptation methods, including VPT~\cite{vpt}, FVP~\cite{fvp}, A2XP~\cite{a2xp}, and VPAD~\cite{vpad}. For fair comparison, we apply a supervised loss to FVP, which was originally proposed in an unsupervised setting. ``Source Only” indicates performance without any adaptation; therefore, even seen domains are treated as unseen.

Across all domains, APEX consistently outperforms existing VP-based methods. On seen domains, it achieves substantial gains over both the baseline and previous methods. As shown in Table~\ref{tab:polyp}, APEX significantly enhances segmentation performance when applied to PraNet~\cite{fan2020pranet}, improving the DICE score by 2.70\% on Domain A and 1.93\% on Domain B compared to baseline (Source Only). Furthermore, it surpasses the recently proposed VPAD method by 1.91\% and 0.70\% on the same domains, respectively.

For unseen domains, prior methods rely on a single prompt optimized on seen distributions, often leading to limited or negligible performance gains. In contrast, APEX consistently improves the performance in unseen domains. Specifically, it improves DICE by 2.79\% on Domain C and 3.33\% on Domain D compared to the baseline. We further observe that this performance trend holds consistently across different backbones. This highlights the adaptability of our framework across a wide range of backbones.

\noindent\textbf{OC/OD Segmentation:}
To further evaluate the effectiveness of the APEX, we validated on OC/OD segmentation task. We optimized the APEX using the training set of REFUGE dataset and evaluated its performance across four domains: REFUGE\_Test (Domain A) and REFUGE\_Val (Domain B) as seen domains, and Drishti-GS (Domain C) and RIM-ONE-r3 (Domain D) as unseen domains.. Table~\ref{tab:optic} shows the experiment results. As shown in the table, APEX consistently outperforms existing VP-based methods. On seen domains, APEX achieves substantial improvements over both baseline and existing prompting methods. For instance, with UNet in Table \ref{tab:optic}, APEX improves the DICE by 4.84\% on Domain A and 4.89\% on Domain B compared to the baseline (Source Only), and surpasses VPAD by 3.43\% and 3.03\% on the same domains, respectively.

On unseen domains, APEX also demonstrates strong generalization. Compared to the baseline on UNet in Table \ref{tab:optic}, it improves DICE by 17.26\% on Domain C and 41.44\% on Domain D. Furthermore, it outperforms VPAD by 15.33\% and 36.82\% on Domains C and D, respectively. Similar results can be observed with other backbones.

\begin{figure}[!t]
    \centering
        \centering
        \includegraphics[width=0.85\linewidth]{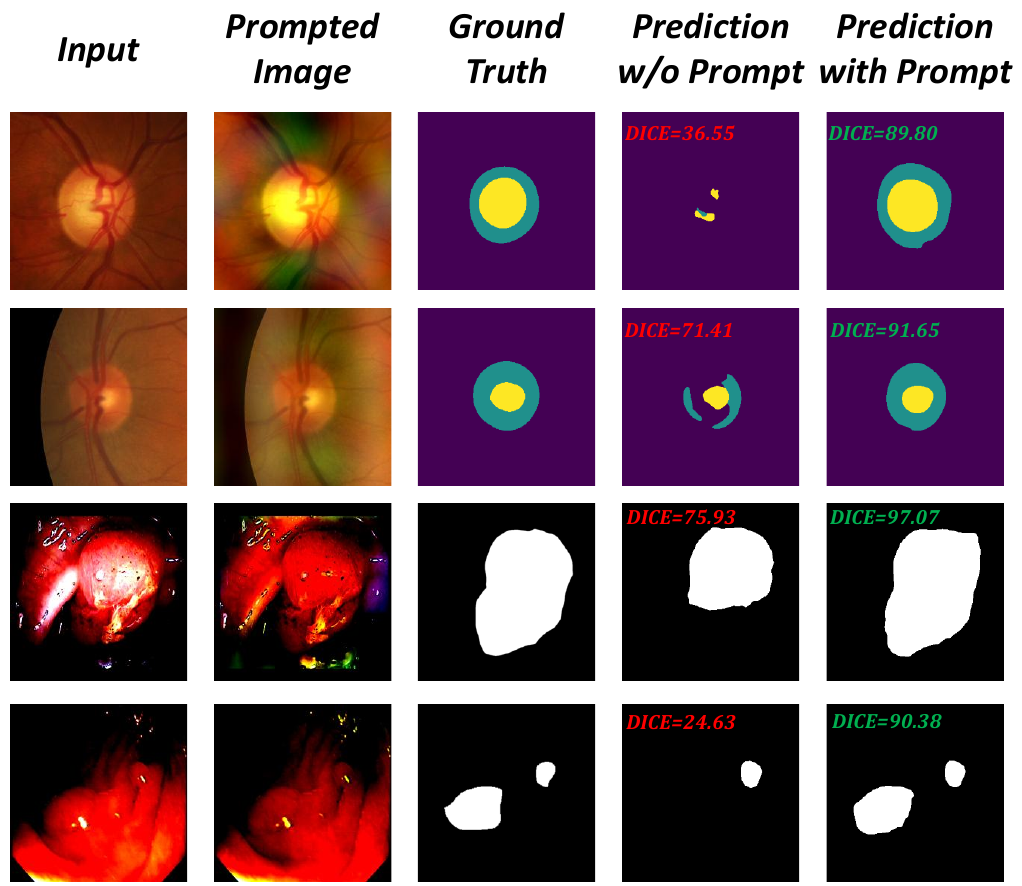}
        \caption{Qualitative comparison between before and after prompting. *In case of polyp images, we visualized normalized images since we prompt into the normalized images with ImageNet statistics.}
        \label{fig:qualitative}
\end{figure}

\noindent\textbf{Experimental Results Summary:}
In summary, experimental results underscore the limitations of existing methods that rely on a single, fixed prompt, which fail to account for intra-domain variability. In contrast, our approach effectively addresses intra-domain variability by dynamically adapting prompts, leading to better adaptability. Furthermore, APEX has strong generalizability to unseen domains without additional optimization. 
In addition, we visualize the results in Fig. \ref{fig:qualitative}. As shown in the figure, without prompting, the model often suffers from incomplete segmentations. However, by applying prompts, the model could make complete segmentations.

\begin{table}[t]
\centering
\resizebox{0.95\linewidth}{!}{%
\begin{tabular}{ccccc}
\specialrule{.1em}{.0em}{0.em}
\textbf{Dataset}          & \textbf{\begin{tabular}[c]{@{}c@{}}Prompt\\ Memory\end{tabular}} & \textbf{\begin{tabular}[c]{@{}c@{}}Avg\\ Seen Domain\end{tabular}} & \textbf{\begin{tabular}[c]{@{}c@{}}Avg\\ Unseen Domain\end{tabular}} & \textbf{\begin{tabular}[c]{@{}c@{}}Avg\\ Total\end{tabular}}   \\ \hline
\multirow{2}{*}{Polyp} & \xmark                                                                & 81.11                                                          & 57.67                                                            & 69.29          \\
                       & \cmark                                                                & \textbf{84.61}                                                 & \textbf{58.27}                                                   & \textbf{70.95} \\ \hline
\multirow{2}{*}{OC/OD} & \xmark                                                                & 86.28                                                          & 76.25                                                            & 83.43          \\
                       & \cmark                                                                & \textbf{87.40}                                                 & \textbf{81.78}                                                   & \textbf{85.81} \\ \specialrule{.1em}{.0em}{0.em}
\end{tabular}}
\caption{Ablation study on the prompt memory in APEX. Dice scores are reported.}
\label{tab:ab_1}
\vspace{-0.5cm}
\end{table}

\subsection{Ablation Study}
\subsubsection{Effect of Prompt Memory:} Table \ref{tab:ab_1} presents an ablation study evaluating the impact of the prompt memory. We used PraNet for the polyp segmentation task and UNet for the OC/OD segmentation task. As shown in the table, introducing the prompt memory leads to consistent performance improvements across all domains. On the polyp segmentation task, it improves the Dice score from 81.11\% to 84.61\% on seen domains and from 57.67\% to 58.27\% on unseen domains. It can be interpreted that the prompt memory effectively enhances both domain-specific adaptability and generalizability, particularly improving performance in unseen domains. Similar results are confirmed in the OC/OD segmentation task.

\subsubsection{Effect of Low-frequency Contrastive Loss:} Table \ref{tab:ab_2} presents an ablation study evaluating the impact of $\mathcal{L}_{LFC}$. As shown in the table, adding $\mathcal{L}_{LFC}$ consistently improves segmentation performance, especially on unseen domains. In the polyp task, it improves the Dice score on unseen domains from 56.83\% to 58.27\%, highlighting its effectiveness in enhancing generalizability to unseen domain. The overall average also improves from 69.20\% to 70.95\%. Similar gains are observed in the polyp segmentation task, confirming the benefit of enforcing domain-discriminative feature learning through LFC framework.

\subsubsection{Effect of Number of Memory Slots:} 
\begin{figure}[t]
    \centering

        \centering
        \includegraphics[width=0.85\linewidth]{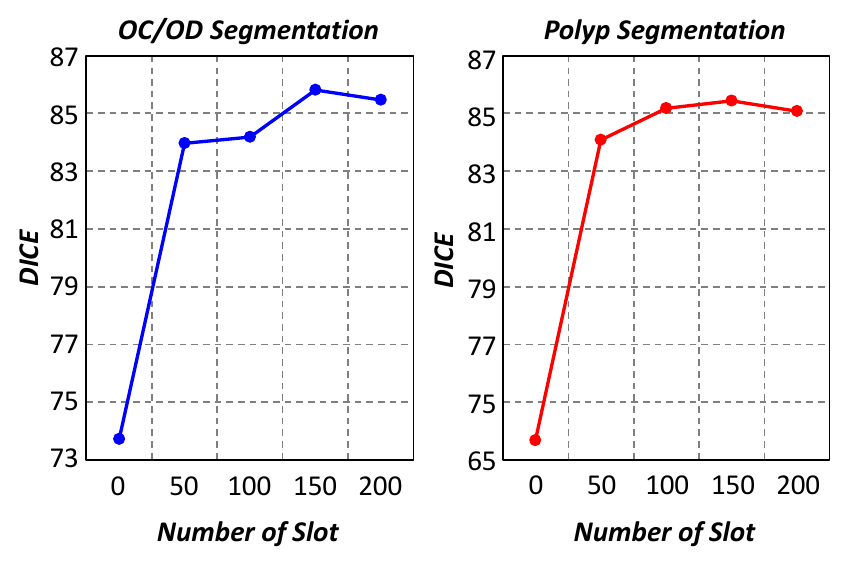}
        \caption{Performance variation according to the number of slots in the Prompt Memory.}
        \label{fig:slot}
        \vspace{-0.5cm}
\end{figure}

We conducted ablation experiments about the impact of the number of slots in the prompt memory. Fig. \ref{fig:slot} shows the experiment results, demonstrating that DICE improves as the number of slots increases. However, when the number exceeds 150, performance is saturated. This is because an excessive number of slots leads to a uniform distribution of the addressing vector, limiting the ability of the memory to extract diverse prompts.

\begin{table}[t]
\centering
\resizebox{0.95\linewidth}{!}{%
\begin{tabular}{ccccc}
\specialrule{.1em}{.0em}{0.em}
\textbf{Dataset}          & \textbf{$\mathcal{L}_{LFC}$} & \textbf{\begin{tabular}[c]{@{}c@{}}Avg\\ Seen Domain\end{tabular}} & \textbf{\begin{tabular}[c]{@{}c@{}}Avg\\ Unseen Domain\end{tabular}} & \textbf{\begin{tabular}[c]{@{}c@{}}Avg\\ Total\end{tabular}}   \\ \hline
\multirow{2}{*}{Polyp} & \xmark             & 83.33                                                          & 56.83                                                            & 69.20          \\
                       & \cmark             & \textbf{84.61}                                                 & \textbf{58.27}                                                   & \textbf{70.95} \\ \hline
\multirow{2}{*}{OC/OD} & \xmark             & 86.71                                                          & 77.77                                                            & 84.17          \\
                       & \cmark             & \textbf{87.40}                                                 & \textbf{81.78}                                                   & \textbf{85.81} \\ \specialrule{.1em}{.0em}{0.em}
\end{tabular}}
\caption{Ablation study on the $\mathcal{L}_{LFC}$ loss. Dice scores are reported.}
\label{tab:ab_2}
\vspace{-0.5cm}
\end{table}

\begin{figure}[t]
    \centering

        \centering
        \includegraphics[width=0.65\linewidth]{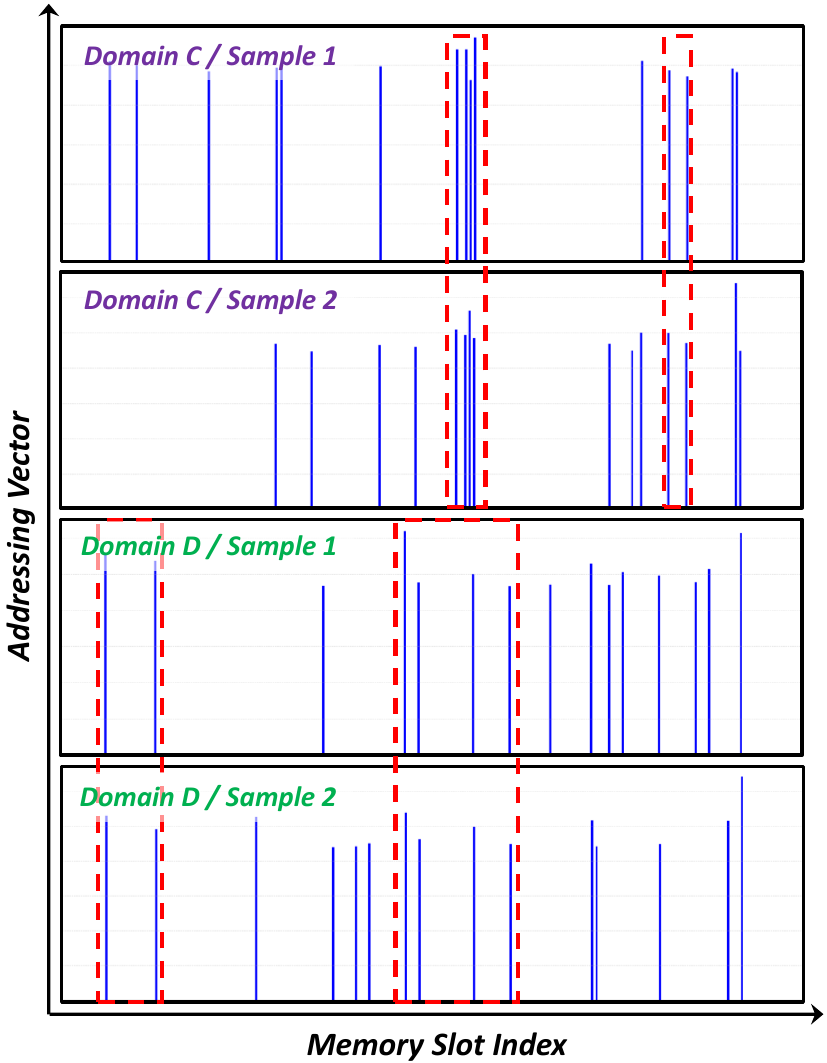}
        \caption{Visualization of the top 10\% activated memory slots for unseen domain samples on the OC/OD segmentation dataset.}
        \label{fig:mem_vis}
        \vspace{-0.3cm}
\end{figure}

\subsubsection{Visualization of Activated Memory Slot:}
To better understand how the prompt memory is utilized in our framework, we visualized the activated memory slots for individual samples on unseen domains. For each sample, we highlighted the top 10\% of memory slots with the highest activation scores. As shown in Fig.~\ref{fig:mem_vis}, the red-dotted boxes indicate commonly activated slots within the same domain. From the figure, we find two key observations:
(1) Within the same domain, certain memory slots are consistently activated across samples, suggesting that the memory captures shared domain-level patterns while preserving instance-specific variability.
(2) Across different domains, distinct memory slots are activated, reflecting the capacity of the memory to encode domain-discriminative features.

These findings demonstrate that APEX adaptively retrieves input-specific prompts by leveraging the discriminative structure of the memory. This adaptive behavior is essential for effectively addressing both intra- and inter-domain variability.

\subsection{Discussion}
\noindent\textbf{Computational Cost:} In the case of the number of parameters and inference time, APEX module requires 0.036M parameters and 20 ms additional time, which is negligible compared to the backbone models, PraNet (32.54M / 800 ms). Further detailed analyses are described in the supplementary material.

\noindent\textbf{Comparison with Test-Time Adaptation:}
When handling data from unseen domains, two common approaches are domain generalization and test-time adaptation (TTA). To compare their practical utility, we evaluate our method (APEX) against VPTTA \cite{vptta}, a TTA-based visual prompting method. As shown in Table~\ref{tab:vptta}, we perform the comparison on unseen domains (Domain C and D) using the PraNet backbone for polyp segmentation. VPTTA adapts the prompt at inference by optimizing on each test image, whereas APEX uses pre-optimized, input-specific prompts derived from seen domains (Domain A and B) without any further optimization.

Although VPTTA achieves slightly higher accuracy by directly optimizing on the test data, this comes at a significant computational cost. Specifically, VPTTA requires approximately 3000 ms per image on a CPU due to its per-sample optimization, while APEX performs inference in just 20 ms—yielding over a 150× speedup. This makes TTA limited to time-sensitive clinical environments, where rapid and reliable inference is crucial.

Moreover, TTA-based methods are known to be sensitive to adaptation order and suffer from performance degradation when repeatedly applied, as errors can accumulate during adaptation \cite{press2023rdumb}. In contrast, APEX offers a robust, plug-and-play solution that generalizes well across domains without incurring additional computational overhead or risking adaptation instability. These results underscore the practicality of our approach for real-world deployment.

\begin{table}[t]
\centering
\resizebox{0.99\linewidth}{!}{%
\begin{tabular}{cccccccc}
\hline
\multirow{2}{*}{\textbf{Backbone}} & \multirow{2}{*}{\textbf{Method}} & \multicolumn{2}{c}{\textbf{Domain C}} & \multicolumn{2}{c}{\textbf{Domain D}} & \multicolumn{2}{c}{\textbf{Avg}} \\ \cline{3-8} 
                                   &                                  & \textbf{DICE}      & \textbf{IoU}     & \textbf{DICE}      & \textbf{IoU}     & \textbf{DICE}   & \textbf{IoU}   \\ \hline
\multirow{2}{*}{\textbf{PraNet}}            & VPTTA                            & 53.82              & 48.34            & \textbf{61.71}              & \textbf{55.04}            & \textbf{60.89}           & \textbf{54.34}          \\
                                   & APEX                             & \textbf{54.43}              & \textbf{48.93}            & 60.85              & 54.91            & 60.18           & 54.28          \\ \hline
\end{tabular}}
\caption{Comparison between test-time adaptation (VPTTA) and domain-generalizable prompting (APEX) on unseen domains for polyp segmentation.}
\label{tab:vptta}
\vspace{-0.3cm}
\end{table}

\noindent\textbf{Future Work:} Our current framework focuses on handling domain shifts caused by different acquisition devices or settings when segmenting the same anatomy. However, clinical applications may also require adaptation across different anatomical structures. In future work, we aim to extend our approach to handle such structural shifts. For this, prompt-based adaptations in the phase or high-frequency components could be a promising direction, as they capture fine-grained and structural information.

\section{Conclusion}
In this work, we introduced APEX, an adaptive visual prompting framework designed to improve the generalizability of prompt-based domain adaptation in medical image segmentation. Unlike existing methods that apply a single prompt uniformly across a domain, APEX adaptively generates input-specific prompts by leveraging a learned memory of domain-discriminative features. To further enhance domain-level feature discrimination, we proposed an LFC learning framework that encourages robust intra-domain clustering and inter-domain discrimination. Extensive experiments on two medical segmentation tasks demonstrated that APEX consistently outperforms SOTA prompting methods on both seen and unseen domains. These results highlight the potential of input-aware, memory-guided prompting as a practical solution for robust medical domain adaptation.
\section*{Acknowledgements}
This work was partly supported by the NRF grant funded by the Korea government(MSIT) (RS-2025-00563942, 30\%), the IITP grant funded by the Korea government(MSIT)(IITP-2026-RS-2020-II201819, 10\%), and the IITP-ITRC grant funded by the Korea government (MSIT)(IITP-2026-RS-2023-00258649, 50\%). Also, this work was supported by the Bayerische Forschungsstiftung (BFS) and the Munich Center for Machine Learning (MCML) during Hong Joo Lee's postdoctoral period. 

{
    \small
    \bibliographystyle{ieeenat_fullname}
    \bibliography{main}
}


\end{document}